\documentclass[10pt,journal,compsoc]{IEEEtran}
\usepackage{lscape}

\ifCLASSOPTIONcompsoc
  \usepackage[nocompress]{cite}
\else
  \usepackage{cite}
\fi
\ifCLASSINFOpdf
\usepackage[pdftex]{graphicx}
\else
\fi
\usepackage{amsmath}
\usepackage{amsfonts}
\relax 
\begin{document}
%
\title{Joint Estimation of Image Representations and their Lie Invariants}
%
%
%
%

\author{Christine~Allen-Blanchette,~\IEEEmembership{Member,~IEEE,}
        and~Kostas~Daniilidis,~\IEEEmembership{Fellow,~IEEE}
}

\IEEEtitleabstractindextext{%
\begin{abstract}
Images encode both the state of the world and its content. The former is useful for tasks such as planning and control, and the latter for classification. The automatic extraction of this information is challenging because of the high-dimensionality and entangled encoding inherent to the image representation. This article introduces two theoretical approaches aimed at the resolution of these challenges. The approaches allow for the interpolation and extrapolation of images from an image sequence by joint estimation of the image representation and the generators of the sequence dynamics. In the first approach, the image representations are learned using probabilistic PCA \cite{tipping1999probabilistic}. The linear-Gaussian conditional distributions allow for a closed form analytical description of the latent distributions but assumes the underlying image manifold is a linear subspace. In the second approach, the image representations are learned using probabilistic nonlinear PCA which relieves the linear manifold assumption at the cost of requiring a variational approximation of the latent distributions. In both approaches, the underlying dynamics of the image sequence are modelled explicitly to disentangle them from the image representations. The dynamics themselves are modelled with Lie group structure which enforces the desirable properties of smoothness and composability of inter-image transformations.
\end{abstract}

}

\maketitle

\IEEEdisplaynontitleabstractindextext

%
\IEEEpeerreviewmaketitle


%
%
%
%

\IEEEraisesectionheading{\section{Introduction}
\label{sec:learing_lie_generators_intro}}
\IEEEPARstart{T}{he} problem of disentangled representation learning has received considerable attention in recent years (\cite{whitney2019disentangling}, \cite{achille2017emergence}, \cite{burgess2018understanding},\cite{mao2017least}).
In robotics, the motivating hypothesis is that a disentangled representation --
a representation for which the semantic information and factors of variation present in disparate ways (\cite{higgins2018towards}, \cite{bengio2012representation})  --
is likely to improve outcomes in planning and control, and classification.
In planning and control, the utility of a representation depends on its ability to describe the world's state independently of
the environmental content and in classification, the utility of a representation depends on its invariance to environmental variation.

Images encode both the state of the world and its content; however, their
high-dimensionality and entangled encoding are limiting. 
The manifold hypothesis argues for the existence of a low-dimensional representation of image data (\cite{bengio2012representation}) and empirical evidence validates this hypothesis. In general, however, an embedded image is not disentangled. Without an explicit mechanism for disentangling semantic information and factors of variation they are likely to remain entangled in the low-dimensional embedding.

This article introduces two approaches for learning disentangled image representations.
In each approach an image sequence is given and
the image representations
and transition dynamics are estimated jointly.
The approaches rely on an assumption that transformations between sequential images are well modelled by a linear Lie group
and that sequential images are close in the transformation space.
The Lie group assumption is well founded since, in general, transformations between sequential images have the appearance of being both smooth and invertible.
When in addition, sequential images are close in the transformation space, the transformations between them are well approximated by the linear combination of transformation generators.

The assumptions above yield two benefits. Firstly, since the image representations and factors of variation are modelled separately, they are disentangled. Secondly, since transformations between sequential images are modelled as the linear combination of Lie generators, extrapolation of an image sequence can be achieved through extrapolation of combination coefficients. The latter is particularly useful for image prediction since extrapolation in $\mathbb{R}^n$ is straight forward whereas extrapolation in the image domain is not.

The approaches introduced in this article build upon the work of \cite{miao2007learning}. The work presented there assumes the underlying transition dynamics of an image sequence are well modelled by a Lie group transformation. The authors use an expectation maximization (EM) framework to estimate the Lie transformation generators and their combination coefficients.
The Lie generators are estimated as parameters and the combination coefficients are estimated as latent variables. The work assumes the availability of a low-dimensional image representation and requires an orthogonalization step in each EM iteration to encourage minimality in the span of the Lie generators.

The two approaches presented in this article address the assumed availability of a low-dimensional image representation through joint estimation of the low-dimensional image representations and transformation generators. In the first approach the low-dimensional image representation is approximated using probabilistic principal component analysis (PCA) \cite{tipping1999probabilistic}. Expectations with respect to the latent posterior distribution have an analytic form, however, the assumed linearity of the underlying image manifold is limiting. The second approach addresses this limitation using probabilistic nonlinear PCA (NPCA). The consequence, however, is an analytically intractable posterior distribution for which a variational approximation is used. 

The article is organized as follows. In Section \ref{sec:op_estim} the approach
of \cite{miao2007learning} is reviewed for its relevance. The next two sections present approaches for joint estimation of the image representations and the transformation generators. In Section \ref{sec:im_op_estim} the image representation is approximated using probabilistic PCA \cite{tipping1999probabilistic}, and in Section \ref{sec:nonlin_im_op_estim} the image representation is approximated using probabilistic NPCA. Section \ref{sec:learning_lie_generators_related_work} gives related work and derivations and proofs are given in the Appendix.

\section{Estimating transition dynamics}
\label{sec:op_estim}
\label{sec:learning_lie_generators_operator_estim}
\begin{figure}[t]
\centering
\vspace{4mm}
\includegraphics[width=0.5\columnwidth]{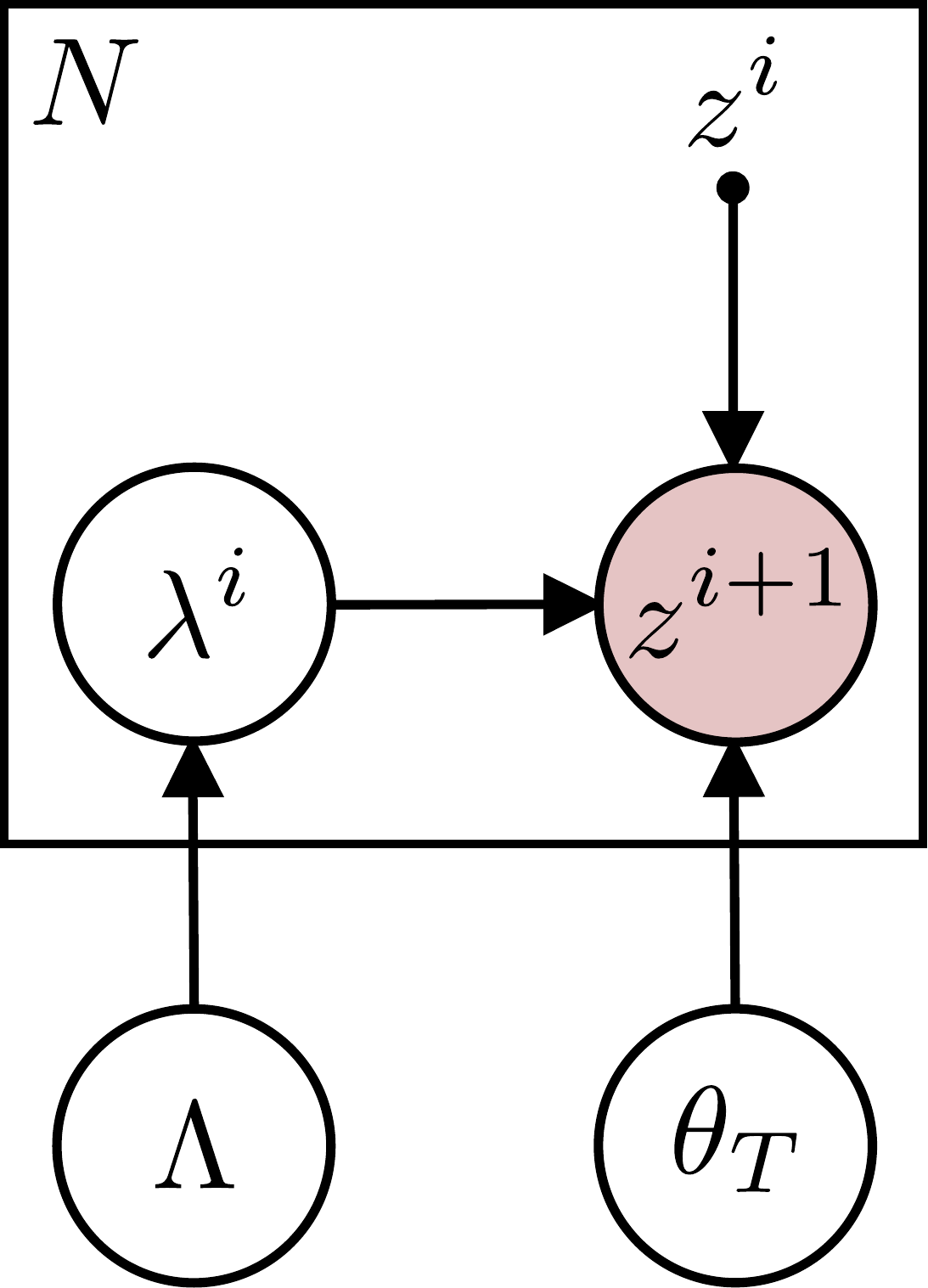}
\caption{Graphical model for estimating transition dynamics \cite{miao2007learning}. The input to the model is the image representation $z^i$. The estimated quantities are: the combination coefficients ($\lambda^i$), the covariance of the combination coefficients ($\Lambda$), the transformation parameters ($\theta_T = (G,\Omega)$) and the transformed image representation ($z^{i+1}$). The transformed image representation $z^{i+1}$ is observed and the number of samples in the dataset is $N$.}
\label{fig:estimating_transition_dynamics}
\end{figure}

The starting point for the work presented in this article is \cite{miao2007learning}. 
There the authors estimate the Lie generators of visual invariance
from image sequences in an EM framework.
The image representations themselves are given
and PCA is used after each EM iteration to ensure the
Lie generators are orthogonal. Because of its relevance, the details of the approach are given here.

In \cite{miao2007learning}, the underlying dynamics of image sequences are modelled by linear Lie groups for their continuity
and group properties. A pair of sequential image representations $\{z^i, z^{i+1}\}$ (in vector form) is assumed to be related by the invertible matrix transformation $H$. In particular the initial image representation $z^i$ is
transformed to the target image representation $z^{i+1}$ by,
\begin{equation}
  \label{eq:op_estim_trans_im}
  z^{i+1} = H z^i.
\end{equation}
Since the transformation $H$ is a Lie group transformation it can be expressed as the matrix exponential of a linear combination of elements $G^j$ in
the Lie algebra, that is, the space of Lie generators,
\begin{equation}
  \label{eq:op_estim_exp_map}
  H = e^{\sum_j \lambda_j G^j}.
\end{equation}
Combining equation (\ref{eq:op_estim_trans_im}) which describes the relationship between pairs of image representations and equation (\ref{eq:op_estim_exp_map}) which relates elements of the Lie group to elements of the Lie algebra gives,
\begin{equation}
  \label{eq:op_estim_trans_im2}
z^{i+1} = e^{\sum_j \lambda^i_j G^j}z^i.
\end{equation}
When the transformation $H$ is small, equation (\ref{eq:op_estim_exp_map}) is well approximated by its first-order approximation. Using the first-order approximation of equation (\ref{eq:op_estim_exp_map}) in equation (\ref{eq:op_estim_trans_im}) gives
\begin{equation*}
z^{i+1} = z^i + {\sum_j \lambda^i_j G^j}z^i.
\end{equation*}
A probabilistic framing is given by the transformation equation
\begin{equation}
  \label{eq:op_estim_prob_transformed_im}
z^{i+1} = z^i + {\sum_j \lambda^i_j G^j}z^i + \epsilon
\end{equation}
where $\epsilon$ is drawn from the Gaussian distribution $\mathcal{N}(0,\Omega)$,
and $\lambda^i$ is drawn from the Gaussian distribution $\mathcal{N}(0,\Lambda)$. In this formulation, the elements $G^j$ of the Lie algebra are the maximum likelihood solutions of a latent variable model.

The likelihood of a transformed image $z^{i+1}$ 
is given by the marginal density
\begin{equation}
  \label{eq:op_estim_prob_transformed_im2}
  p(z^{i+1} | z^i, \theta_T, \Lambda) = \int p(z^{i+1} | z^i, \lambda^i, \theta_T ) p(\lambda^i | \Lambda) d\lambda^i, 
\end{equation}
where $\theta_T=(G, \Omega)$ are the transformation parameters.

Due to the integral in the marginal, 
setting derivatives of the log-likelihood function to zero
does not give a closed form solution. A common method for
finding maximum likelihood solutions in this setting is to use the expectation maximization (EM) algorithm.

EM is an alternating
estimation framework in which 
the likelihood of the joint distribution of the data
and latent variables (complete-data likelihood),
is maximized by alternating between
estimation of the latent variables 
and the distribution parameters.
These are referred to as the expectation-step (E-step) 
and maximization-step (M-step) respectively.
The alternating estimation strategy is guaranteed to converge to a local optimum under iteration.

The complete-data likelihood of the transformed image representation $z^{i+1}$ and the
combination coefficients $\lambda^i$ is given by the Gaussian distribution,
\begin{equation}
  \label{eq:op_estim_complete-data_td}
\prod_i p(\lambda^i, z^{i+1} | z^i, \theta_T, \Lambda) = \prod_i p(z^{i+1} | z^i, \lambda^i , \theta_T, \Lambda)\, p(\lambda^i| \Lambda)
\end{equation}
(see Section \ref{sec:complete_data_op} for derivation details). The E-step is performed with respect to the latent variables $\lambda^i$
and the M-step is performed with respect to the distribution parameters
$G$ and $\Omega$.

Because a prior is given for the combination coefficients $\lambda^i$,
their estimates are given by maximum a posteriori (MAP) likelihood estimate.
The MAP estimate is the expected value of the
posterior distribution of the combination coefficients. The posterior distribution of the combination coefficients $\lambda^i$ is given by,
\begin{equation*}
  \label{eq:op_estim_latent_posterior_td}
  p(\lambda^i | z^{i+1}, z^i, \theta_T,\Lambda) = \mathcal{N}(\lambda^i | q, K )
\end{equation*}
where
\begin{equation*}
  K = (\Lambda^{-1} + A^T\Omega^{-1} A)^{-1}, \quad q = K A^T\Omega^{-1}\Delta z^i,
\end{equation*}
and
\begin{equation}
\label{eq:op_estim_A_def}
  A_{\cdot,m} = G^m z^i
\end{equation}
where $\Delta z^i= z^{i+1} - z^i$ is the difference between sequential image representations (see Section \ref{sec:latent_posterior_op} for derivation details).

Estimates for the distribution parameters $G$ and $\Omega$ are
given by their maximum likelihood (MLE) solutions.
The MLE solutions are found by 
computing partial derivatives of the log-likelihood function
and solving for parameter values at the critical point. The resulting update equations for $G$ and $\Omega$ are
\begin{equation*}
G = \left(\sum_i \Delta z^i(z^i\otimes\mathbb{E}[\lambda^i])^T\right)\left(\sum_i z^i(z^i)^T\otimes \mathbb{E}[\lambda^i (\lambda^i)^T]\right)^{-1}
\end{equation*}
and
\begin{align*}
  \Omega = \frac{1}{M}\Big( &\sum_i \Delta z^i(\Delta z^i)^T - 2A\mathbb{E}[\lambda^i](\Delta z^i)^T + A\mathbb{E}[\lambda^i (\lambda^i)^T]A^T\Big)
\end{align*}
with $A$ defined as in equation (\ref{eq:op_estim_A_def}) (see Section \ref{sec:optim_estim_m_step} for derivation details).

Although convergence of EM is guaranteed, the learned generators $G$ may not be orthogonal; what's more, in general the number of transformations $G^j$ is unknown. To remedy this, each iteration of EM is followed by an orthogonalization step. Application of PCA to $G$ resolves these issues by providing a minimal set of orthogonal $G^j$ to be used in the next iteration of EM.

A graphical model of the approach described in this section is given in Figure \ref{fig:estimating_transition_dynamics}. The input to the model is the image representation $z^i$. The estimated quantities are: the combination coefficients of the Lie generators ($\lambda^i$), the covariance of the combination coefficients ($\Lambda$), the transformation parameters ($\theta_T = (G,\Omega)$, the Lie generators and the transformation covariance) and the transformed image representation ($z^{i+1}$). The transformed image representation $z^{i+1}$ is observed and the number of samples in the dataset is $N$.

\section{Joint estimation of PPCA image representations and transition dynamics}
\label{sec:im_op_estim}
\label{sec:learning_lie_generators_image_op_estim}
\begin{figure}[t]
\centering
\vspace{4mm}
\includegraphics[width=0.7\columnwidth]{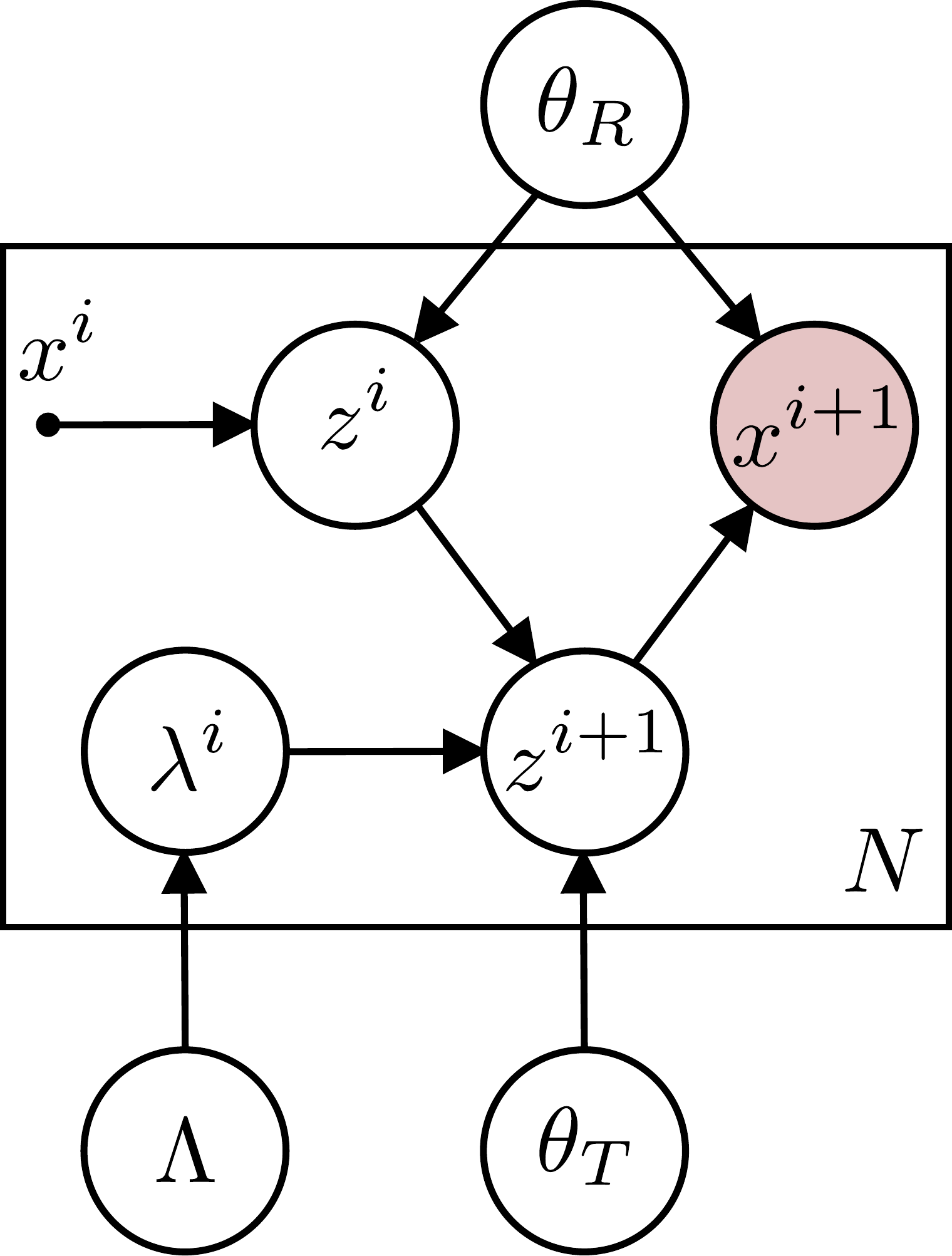}
\caption{Graphical model for joint estimation of PPCA \cite{tipping1999probabilistic} image representations and transition dynamics. The input to the model is the image $x^i$. The estimated quantities are: the image representations ($z^i$, $z^{i+1}$), the combination coefficients ($\lambda^i$), the covariance of the combination coefficients ($\Lambda$), the transformation parameters ($\theta_T = (G,\Omega)$), the reconstruction parameters ($\theta_R=(W, \mu, \sigma)$), and the transformed image ($x^{i+1}$). The transformed image $x^{i+1}$ is observed and the number of samples in the dataset is $N$.}
\label{fig:PPCA_transition_dynamics}
\end{figure}

This and the following section introduce approaches which extend the work of \cite{miao2007learning} to jointly estimate Lie transition dynamics and low-dimensional image representations.
In \cite{miao2007learning}, the transition dynamics of an image sequence are estimated in an EM framework. The transformations between sequential images are assumed to be well modelled by a linear Lie group. The Lie generators are estimated as parameters of the model and the combination coefficients are estimated as latent variables. The image representations themselves are assumed to be given. 

In this section the transition dynamics are assumed to be well modelled by a linear Lie group and the images are assumed to be well approximated by a low-dimensional linear subspace modelled with probabilistic principal component analysis (PPCA) \cite{tipping1999probabilistic}.
The transition dynamics and image representations are estimated jointly in an EM framework.

Principal component analysis (PCA) estimates a d-dimensional least squares best-fit linear subspace in which to represent the data.
When framed probabilistically, the low-dimensional image representation is the maximum likelihood solution of a latent variable model.
An image $x^k$ is expressed in terms of a low-dimensional embedding (or latent variable) $z^k$ by 
\begin{equation*}
  x^k = Wz^k + \mu + \epsilon
\end{equation*}
where $\epsilon$ is isotropic Gaussian noise with variance $\sigma^2$ and the embedding $z^k$ is drawn from a standard-normal distribution.

The likelihood of the transformed data $x^{i+1}$ is given by the marginal density
\begin{align*}
&p(x^{i+1} | x^i, \theta_T, \Lambda, \theta_R) = \\&\int p(x^{i+1} | z^{i+1}, \theta_R) \,p(z^{i+1} | z^i, \theta_T, \Lambda)\, p(z^{i} | x^{i}, \theta_R)\, dz^{i+1}\, dz^i
\end{align*}
where $\theta_T=(G, \Omega)$ are the transformation parameters and $\theta_R=( W, \mu, \sigma)$ are the representation embedding parameters. The probability of the transformed latent representation $z^{i+1}$ is given by equation (\ref{eq:op_estim_prob_transformed_im2}). It is repeated here for the reader's convenience,
\begin{equation*}
  p(z^{i+1} | z^i, \theta_T, \Lambda) = \int p(z^{i+1} | z^i, \lambda^i, \theta_T ) p(\lambda^i | \Lambda) d\lambda^i.
\end{equation*}

The posterior distribution of the latent image representation given the image data is 
\begin{equation}
  \label{eq:im_op_im_rep_posterior}
  p(z^i|x^i, \theta_R) = \mathcal{N}(z^i|u, \Sigma)
\end{equation}
where
\begin{equation*}
  \Sigma = I + \sigma^{-2}W^TW, \quad u = \sigma^{2}\Sigma^{-1} W^T x^i_\mu
\end{equation*}
where $x^i_\mu = x^i - \mu$ is $x^i$ mean-centered
(see Section \ref{sec:posterior_image} for derivation details).

Due to the integral in the marginal, setting derivatives of the log likelihood function to zero
does not give a closed form solution. The problem does however yield itself to estimation using EM. The complete-data likelihood is given by,
\begin{align*}
  \prod_i& p(\lambda^i, z^{i+1}, x^{i+1}, z^i| x^i, \theta_T, \Lambda, \theta_R) = \\&\prod_i p(x^{i+1} | z^{i+1}, \theta_R)\, p(z^{i+1} | z^i, \lambda^i , \theta_T, \Lambda)\, p(\lambda^i| \Lambda) \,p(z^i | x^i, \theta_R)
\end{align*}

Because a prior is given for the latent representations of the initial images $z^i$, target images, $z^{i+1}$, and the combination coefficients $\lambda^i$ of the Lie generators, their estimates are given by
maximum a posteriori (MAP) likelihood estimation. The MAP estimate is the expected
value of the posterior distribution of the latent variables $z^i$, $z^{i+1}$ and $\lambda^i$. The posterior distribution is given by,
\begin{align}
  \label{eq:latent_posterior}
  p(&\lambda^i, z^{i+1}, z^i | x^i, x^{i+1}, \theta_T, \Lambda, \theta_R) = \\& p( z^{i+1} | x^{i+1}, z^i, \lambda^i, \theta_T, \theta_R )\, p(\lambda^i| \Lambda)\, p(z^i | x^i, \theta_R)
\end{align}
where
\begin{align*}
  p(& z^{i+1} | x^{i+1}, z^i, \lambda^i, \theta_T, \theta_R ) = \\&\frac{ p(x^{i+1} | z^{i+1}, \theta_R)\,p(z^{i+1} | z^i, \lambda^i, \theta_T) }{p(x^{i+1} |\theta_R)} = \mathcal{N}(z^{i+1}| \gamma, \Gamma)
\end{align*}
with
\begin{align*}
  \gamma = \Gamma \{ \sigma^{-2}W^T x^{i+1}_\mu + \Omega^{-1}(z^i + A\lambda^i)\},\\
  \Gamma = (\Omega^{-1} + \sigma^{-2} W^T W)^{-1}, \quad A_{\cdot,m} = G^m z^i
\end{align*}
(see Section \ref{sec:im_op_estim_latent} for derivation details). The latent posterior distribution in equation (\ref{eq:latent_posterior}) is non-Gaussian; however, it is a member of the exponential family. Consequently, there is an analytic form for its sufficient statistics (see Section \ref{sec:im_op_estim_latent} for derivation details).

Estimates for the distribution parameters $G, \Omega, W, \mu$ and $\sigma$ are given by their maximum likelihood
solutions which require computing partial derivatives of the log likelihood function and
solving for parameter values at the critical point. The resulting update equations for $G, \Omega, W$ and $\sigma$ are
\begin{align*}
G =& \left(\sum_i \mathbb{E}[\Delta z^i(z^i\otimes\lambda^i)^T]\right)\left(\sum_i \mathbb{E}[z^i(z^i)^T \otimes \lambda^i(\lambda^i)^T] \right)^{-1}\\
  \Omega =& \frac{1}{M}\Big( \sum_i \mathbb{E}[\Delta z^i(\Delta z^i)^T] - 2A\mathbb{E}[\lambda^i(\Delta z^i)^T]\\&\qquad\qquad + A\mathbb{E}[\lambda^i (\lambda^i)^T]A^T\Big)\\
  W =& \Big(\sum_i x^{i+1}_\mu \mathbb{E}[z^{i+1}]^T + x^i_\mu \mathbb{E}[z^i]^T\Big)\\&\Big(\sum_i \mathbb{E}[z^{i+1}(z^{i+1})^T] + \mathbb{E}[z^i(z^i)^T]\Big)^{-1},
\end{align*}
and
\begin{align*}
    \sigma^2 =& \frac{1}{ND} \text{tr}\Big(\sum_i x^i_\mu (x^i_\mu)^T + x^{i+1}_\mu(x^{i+1}_\mu)^T \\&\qquad\qquad\quad + W(\mathbb{E}[z^i(z^i)^T] + \mathbb{E}[z^{i+1}(z^{i+1})^T])W^T\Big) \\&-\frac{2}{ND} \text{tr}\left(\sum_i W(\mathbb{E}[z^i] (x^i_\mu)^T + \mathbb{E}[z^{i+1}](x^{i+1}_\mu)^T) - \mu\mu^T\right)
\end{align*}
where $N$ is the number of samples and $D$ is the dimension of $x^k$.
The maximum likelihood estimate for $\mu$ is the average of the data.

Orthogonalization of the Lie generators is required at each iteration of EM.

A graphical model of the approach described in this section is given in
Figure \ref{fig:PPCA_transition_dynamics}. The input to the model is the image $x^i$. The estimated quantities are: the image representations ($z^i$, $z^{i+1}$), the combination coefficients of the Lie generators ($\lambda^i$), the covariance of the combination coefficients ($\Lambda$), the transformation parameters ($\theta_T = (G,\Omega)$, the Lie generators and the transformation covariance), the reconstruction parameters ($\theta_R=(W, \mu, \sigma)$, the low-dimensional linear subspace, image average and covariance), and the transformed image ($x^{i+1}$). The transformed image $x^{i+1}$ is observed and the number of samples in the dataset is $N$.

\section{Joint estimation of PNPCA image representations and transition dynamics}
\label{sec:nonlin_im_op_estim}
\label{sec:learning_lie_generators_nonlin_image_op_estim}
\begin{figure}[t]
\centering
\vspace{4mm}
\includegraphics[width=0.6\columnwidth]{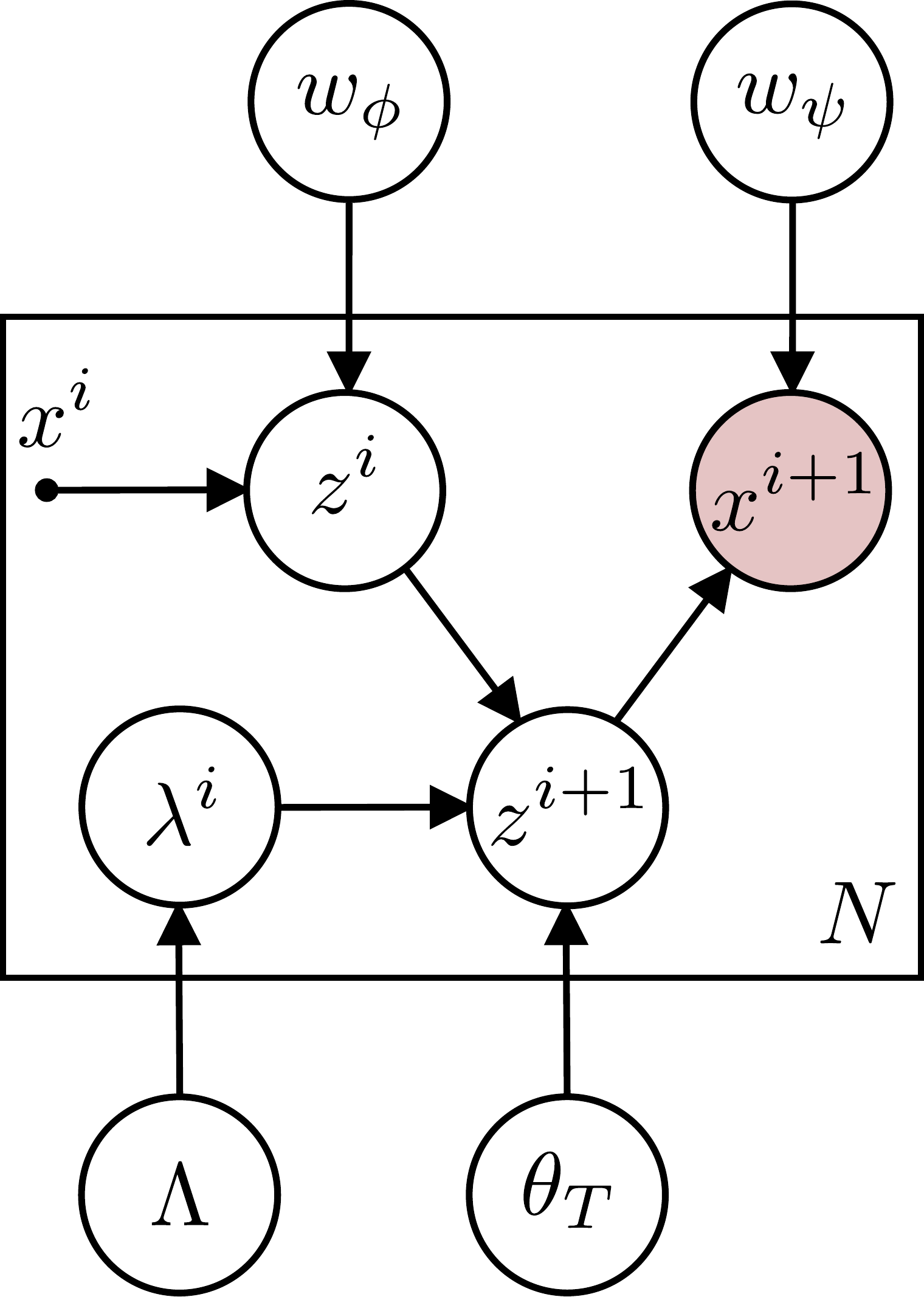}
\caption{Graphical model of joint estimation of PNPCA image representations and transition dynamics. The input to the model is the image $x^i$. The estimated quantities are: the image representations ($z^i$, $z^{i+1}$), the combination coefficients ($\lambda^i$), the covariance of the combination coefficients ($\Lambda$), the transformation parameters ($\theta_T = (G,\Omega)$), the reconstruction parameters ($w_\psi$, $w_\phi$), and the transformed image ($x^{i+1}$). The transformed image $x^{i+1}$ is observed and the number of samples in the dataset is $N$.}
\label{fig:NPCA_transition_dynamics}
\end{figure}

The previous section introduced an approach for joint estimation of transition dynamics and low-dimensional image representations. The approach for estimating the image representation assumed linearity of the low-dimensional image manifold. This assumption is quite restrictive and in this section, the low-dimensional image manifold is allowed to be nonlinear.

Nonlinear PCA (NPCA) can be implemented using an autoencoding network
$f(\cdot)$ defined as the composition of encoding
and decoding networks
$\phi$ and $\psi$,
\begin{equation*}
  f(\cdot) = \psi(\phi(\cdot)).
\end{equation*}
The parameters of the networks $\phi$ and $\psi$
are denoted $w_\phi$ and $w_\psi$.

When framed probabilistically, an image $x^k$ is expressed in terms of a low-dimensional embedding (or latent variable) $z^k$ by,
\begin{equation*}
  x^k = \psi(z^k) + \epsilon
\end{equation*}
where $\epsilon$ is isotropic Gaussian noise with variance $\sigma^2$ and the embedding $z^k$ is drawn from a standard-normal distribution.

The likelihood of the transformed data $x^{i+1}$ is given by the marginal density
\begin{align*}
  \label{eq:marginal}
p(&x^{i+1} | x^i, \theta_T, \Lambda, w_\psi) = \\&\int p(x^{i+1} | z^{i+1}, w_\psi)\, p(z^{i+1} | z^i, \theta_T, \Lambda)\, p(z^{i} | x^{i}, w_\psi) \,dz^{i+1}\, dz^i
\end{align*}
where $\theta_T=(G, \Omega)$ are the transformation parameters and the probability of the transformed latent representation $z^{i+1}$ is given by equation (\ref{eq:op_estim_prob_transformed_im2}). It is repeated here for the reader's convenience,
\begin{equation*}
  p(z^{i+1} | z^i, \theta_T, \Lambda) = \int p(z^{i+1} | z^i, \lambda^i, \theta_T ) p(\lambda^i | \Lambda) d\lambda^i.
\end{equation*}
The posterior distribution of the latent image representation given the image data is analytically intractable because of the nonlinear dependence of $\phi(z^i)$ on $z^i$. A common approach in this setting is to use variational Bayes, an approach in which the analytically intractable distribution $p(z^{i} | x^{i}, w_\psi)$ is approximated by a tractable distribution $q(z^{i} | x^{i}, w_\phi)$. 

The distribution $q(z^{i} | x^{i}, w_\phi)$ is chosen to be Gaussian with the mean and standard deviation of the distribution determined by the encoding network,
\begin{equation*}
q(z^{i} | x^{i}, w_\phi ) = \mathcal{N}(z^i | \phi_\mu(x^i), \phi_\sigma(x^i)).
\end{equation*}
To encourage the variational approximation to be close to the true distribution, a KL-divergence between the true and approximate distributions is introduced into the log-likelihood function giving,
\begin{align*}
  \sum_i &\log p(\lambda^i, z^{i+1}, x^{i+1}, z^i | x^i, \theta_T, \Lambda, w_\psi, w_\phi) \approx \\
  &\sum_i \log p(x^{i+1} | z^{i+1}, w_\psi) + \log p(z^{i+1} | z^i, \lambda^i , \theta_T) \\ &\qquad +\log p(\lambda^i| \Lambda) + 
  \mathcal{L}(q(z|x, w_\phi)),
\end{align*}
where the lower-bound $\mathcal{L}(q(z|x, w_\phi))$ is given by,
\begin{align*}
  \mathcal{L}(q(z|x, w_\phi)) =& - \text{KL}(q(z|x, w_\phi)\,||\,p(z | w_\psi)) \\&+ \mathbb{E}_{q(z|x, w_\phi)}[p(x|z, w_\psi)].
\end{align*}
The approximate posterior of the latent variable distribution is given by,
\begin{align}
  \label{eq:nonlin_op_estim_latent_posterior}
  p(&\lambda^i, z^{i+1}, z^i | x^i, x^{i+1}, \theta_T, \Lambda, w_\psi, w_\phi) \approx \\&  \frac{q(z^{i+1} | x^{i+1})}{p(z^{i+1} | w_\psi)}\, p(z^{i+1} | z^i, \lambda^i, \theta_T)\, p(\lambda^i| \Lambda)\, q(z^i | x^i, w_\phi)
\end{align}
where the standard-Gaussian is chosen for the prior $p(z^{i+1} | w_\psi)$ (see Section \ref{sec:nonlin_op_estim_latent} for derivation details). 

Estimates for the parameters $G, \Omega$, are given by their
maximum likelihood solution which requires computing partials of the
complete-data log likelihood function and solving for parameter values at the critical point. The resulting update equations for 
$G, \Omega$,
are
\begin{equation*}
G = \left(\sum_i \mathbb{E}[\Delta z^i(z^i\otimes\lambda^i)^T]\right)\left(\sum_i \mathbb{E}[z^i(z^i)^T \otimes \lambda^i(\lambda^i)^T] \right)^{-1}
\end{equation*}
and
\begin{equation*}
  \Omega = \frac{1}{M}\Big( \sum_i \mathbb{E}[\Delta z^i(\Delta z^i)^T] - 2A\mathbb{E}[\lambda^i(\Delta z^i)^T] + A\mathbb{E}[\lambda^i (\lambda^i)^T]A^T\Big).
\end{equation*}
The neural network parameters $w_\phi$ and $w_\psi$ are updated
by backpropagation using the reparameterization trick (\cite{kingma2013auto}).

Orthogonalization of the Lie generators is required at each iteration of EM.

A graphical model of the approach described in this section is given in Figure \ref{fig:NPCA_transition_dynamics}.
Graphical model of joint estimation of PNPCA image representations and transition dynamics. The input to the model is the image $x^i$. The estimated quantities are: the image representations ($z^i$, $z^{i+1}$), the combination coefficients of the Lie generators ($\lambda^i$), the covariance of the combination coefficients ($\Lambda$), the transformation parameters ($\theta_T = (G,\Omega)$, the Lie generators and the transformation covariance), the reconstruction parameters ($w_\psi$, $w_\phi$), and the transformed image ($x^{i+1}$). The transformed image $x^{i+1}$ is observed and the number of samples in the dataset is $N$.

\section{Related Work}
\label{sec:learning_lie_generators_related_work}
Each of the two approaches presented in this chapter jointly estimate low-dimensional image representations and Lie transformation generators for a given image sequence. The first approach employs probabilistic principle component analysis (PCA) which assumes linearity of the low-dimensional image manifold. The second approach uses probabilistic nonlinear PCA (NPCA) which allows the image manifold to be nonlinear. An analytic formulation is given for the posterior distribution in each setting; however, in the second the posterior is approximate.

This section describes select related work in modeling the underlying dynamics of, or variation in, the image data.

\subsubsection{Estimating Lie generators}
\cite{freeman1991design, perona1995deformable, teo1998design}, and \cite{bansal2014steerability}
propose techniques for estimating a steering basis on linear Lie groups. 
Steering can be described as transforming a function defined on a group 
by the group action using a linear combination of basis functions and combination coefficients soley dependent on the steering direction.
 The steering basis is determined from a known Lie transformation group and not from
 a sequence of images.

\subsubsection{Probabilistic modeling with Lie dynamics}
\cite{miao2007learning} estimate Lie transition generators and combination coefficients from an image sequence in an EM framework.
The image representations from which
the dynamics are estimated are assumed to be given and sequential images are assumed to be close in the transformation space.

\cite{sohl-dickstein2010unsupervised} use a similar framework to estimate Lie transition generators and their combination coefficients but do not require sequential images to be close in the transformation space. The resulting nonconvexity in inference is addressed using a coarse-to-fine estimation of the transformation generators. The generators themselves are constrained to be diagonalizable and consequently do not capture transformations such as constrast, scaling and translation without periodic boundary. By relaxing the Lie group assumption, however, \cite{sohl-dickstein2010unsupervised} demonstrate how their technique can be used to capture a fuller set of transformations.

\cite{cohen2014learning} introduce an approach
for probabilistic estimation of special orthogonal transition dynamics
from an image sequence. The authors model the transformation coefficients
using the von-Mises distribution and show that the posterior distribution of the transformation coefficients is also von-Mises.

\cite{falorsi2018explorations} introduce an approach for probabilistic estimation of low-dimensional image representations that are compatible with the action of special orthogonal transformations in 3D. Images are mapped to the Fourier domain where they are transformed by a group action. The representation of the group action is given. \cite{falorsi2019reparameterizing} extends \cite{falorsi2018explorations} to accomodate other Lie transformations groups but also requires the group representation to be provided.

\subsubsection{Probabilistic modeling of transition dynamics}
There are many other approaches for estimating
disentangled image representations from video sequences
using deep neural networks. These approaches typically do not
directly model the trasition dynamics in matrix form
and when they do, the dynamics are not constrained to have the Lie group structure.

For example,
\cite{watter2015embed} learn to estimate state dependent
locally linear discrete-time transition dynamics. 
A learned transformation of
an image gives the (low-dimensional) state vector, and
a learned embedding of the state vector gives the
locally linear transition dynamics. During training,
the authors must regularize their loss function 
to ensure estimates in the state space correspond
to embeddings of the observation space.

Another example comes from
\cite{whitney2019disentangling} where latent representations
of varying factors are learned from a video sequence by defining a set of
factors which which evolve in time
and from which reconstruction of an image is possible.
The time evolution of each factor is
determined by an MLP and is conditionally
dependent on its representation at the prevous time step.

\section{Conclusion}
This article introduces two generative models for jointly estimating image representations and their transition dynamics. The transition dynamics are constrained to have the Lie group structure for smoothness and composability of transformations. Because the Lie generators are learned, transformations between image frames can be interpolated by interpolating between combination coefficients in $\mathbb{R}^n$ which, in contrast to interpolation in the image space, is well understood. The approaches differ in the way the images are encoded. 
\appendices
\label{sec:appendix}
\section{Combining Gaussian variables}
\label{sec:combining_gauss}
\subsection{Linear Gaussian variable}
In this section a form for the marginal, conditional and joint distributions
are given for the case where the conditional distribution $p(y|x)$
is a linear Gaussian model, that is, when $p(y|x)$
has a mean that is a linear function of $x$ and a covariance that
is independent of $x$. The material in this section is adapted
primarily from \cite{bishop2006pattern}, details of the derivations are given in Section 2.3 of the same text.

Beginning with the linear Gaussian $p(y|x)$ and $p(x)$,
\begin{align*}
  p(x) &= \mathcal{N}(x|\mu, \Lambda^{-1})\\
  p(y|x) &= \mathcal{N}(y| A x + b, L^{-1}),
\end{align*}
the marginal distribution $p(y)$ and conditional distribution $p(x|y)$
are given by
\begin{align*}
  p(y) &= \mathcal{N}(y| A\mu + b, L^{-1} + A\Lambda^{-1}A^T)\\
  p(x|y) &= \mathcal{N}(x|\Sigma\{A^TL(y-b)+\Lambda\mu\}, \Sigma)
\end{align*}
where
\begin{align*}
  \Sigma =& (\Lambda + A^TLA)^{-1}.
\end{align*}
The joint distribution $p(x,y)$ is given by
\begin{align*}
  p(x,y) &=\mathcal{N}(x, y | m, R^{-1})
\end{align*}
where
\begin{align*}
  R^{-1} =
  \begin{pmatrix}
    \Lambda^{-1} & \Lambda^{-1}A^T\\
    A\Lambda^{-1} & L^{-1} + A\Lambda^{-1}A^T
  \end{pmatrix}, \quad
  m = \begin{pmatrix}
    \mu\\
    A\mu + b
    \end{pmatrix}.
\end{align*}

\subsubsection{Gaussian joint distribution}
In this section the conditional and marginal distributions of two sets of variables are given for the case when their joint distribution is Gaussian. The material
in this section is adapted primarily from \cite{bishop2006pattern}, details of the derivations are given in
Section 2.3 of that text.

For the variable $x\sim\mathcal{N}(\mu, \Lambda^{-1})$ partitioned into two disjoint subsets $x_a$ and
$x_b$ so that
\begin{equation*}
  x = \begin{pmatrix}
    x_a\\
    x_b
  \end{pmatrix}.
\end{equation*}
The corresponding partitions of the mean and precision matrix are given by
\begin{equation*}
  \mu =
  \begin{pmatrix}
    \mu_a\\
    \mu_b
  \end{pmatrix}, \quad
  \Lambda =
  \begin{pmatrix}
    \Lambda_{aa} & \Lambda_{ab}\\
     \Lambda_{ab} & \Lambda_{bb}
  \end{pmatrix}
\end{equation*}
and the conditional distribution $p(x_a|x_b)$ is Gaussian with
sufficient statistics
\begin{equation*}
  \mu_{a|b} = \mu_a - \Lambda_{aa}^{-1}\Lambda_{ab}(x_b - \mu_b), \quad
  \Sigma_{a|b}=\Lambda_{aa}^{-1}.
\end{equation*}

\section{Variational Bayes}
\label{sec:variational_bayes}
Variational Bayes is an approximation technique commonly used when evaluation of the posterior distribution or evaluation of expectations with respect to the posterior distribution is computationally intractable. This can be because of high dimensionality of the posterior distribution or the analytic intractability of its expression.

To resolve the computational expense the posterior distribution $p(z|x)$ 
is approximated with an analytically tractable distribution $q(z|x)$. 
In this setting the log-likelihood of the data is given by,
\begin{equation}
  \log p(x) = \text{KL}(q||p) + \mathcal{L}(q)
\end{equation}
where the first term is the KL-divergence,
\begin{equation*}
   \text{KL}(q||p) = -\int q(z) \log \frac{p(x|z)}{q(z)} dz
\end{equation*}
and the second term is the lower bound,
 \begin{equation*}
   \mathcal{L}(q) = \int q(z) \log \frac{p(x,z)}{q(z)} dz
 \end{equation*}

\section{Derivations for transition dynamics (Sec \ref{sec:op_estim})}
\subsection{Complete-data likelihood}
\label{sec:complete_data_op}
The complete-data likelihood is the product of the data and latent variable distributions,
\begin{align*}
  \prod_i p(\lambda^i, z^{i+1} | z^i, \theta_T, \Lambda) &= \prod_i p(z^{i+1} | z^i, \lambda^i, \theta_T)\, p(\lambda^i | \Lambda)
 \\ &= \prod_i\mathcal{N}(z^{i+1}| z^i + A\lambda^i, \Omega)\, \mathcal{N}(\lambda^i| 0, \Lambda)
\end{align*}
where $\theta_T=(G, \Omega)$ are the transformation parameters.
The form given in
equation (\ref{eq:op_estim_complete-data_td}) is derived using the equations in Section 
\ref{sec:combining_gauss}. 
The transformed image distribution is a linear Gaussian distribution and the prior distribution on the combination coefficients is Gaussian. Using the equations in Section \ref{sec:combining_gauss}, the complete-data likelihood can be expressed
\begin{equation}
  \label{eq:deriv_op_estim_complete_data}
  p(\lambda^i, z^{i+1} | z^i, \theta_T, \Lambda) = \mathcal{N}(\lambda^i, z^{i+1} | m, R^{-1})
\end{equation}
where the parameters $R^{-1}$ and $m$ are given by
\begin{equation*}
  R^{-1} =
  \begin{pmatrix}
    \Lambda & \Lambda A^T\\
    A\Lambda & \Omega + A\Lambda A^T
  \end{pmatrix}, \quad
  m = \begin{pmatrix}
    0\\
    z^i
    \end{pmatrix}
\end{equation*}
and
\begin{equation*}
  A_{\cdot,j} = G^j z^i.
\end{equation*}

\subsection{Posterior distribution of the latent variables}
\label{sec:latent_posterior_op}
The latent posterior distribution of the combination coefficients $\lambda^i$ given in equation (\ref{eq:op_estim_latent_posterior_td}) is derived using the equations in Section \ref{sec:combining_gauss}. 
The transformed image distribution is a linear Gaussian distribution and
the prior distribution on the combination coefficients is Gaussian.
Using the equations in Section \ref{sec:combining_gauss}, the latent posterior distribution on the combination coefficients is given by
\begin{equation*}
  p(\lambda^i | z^{i+1}, z^i, \theta_T,\Lambda) = \mathcal{N}(\lambda | q, K )
\end{equation*}
where $\theta_T=(G, \Omega)$ are the transformation parameters, and where
\begin{equation*}
  K = (\Lambda^{-1} + A^T\Omega^{-1} A)^{-1}, \quad q = K A^T\Omega^{-1}\Delta z^i
\end{equation*}
where $\Delta z_i=z^{i+1} - z^i$ is the difference between sequential image representations, and
\begin{equation*}
  A_{\cdot,j} = G^j z^i.
\end{equation*}

\subsection{Update equations for the model parameters}
\label{sec:optim_estim_m_step}
The update equation for each distribution parameter is found by computing the partial derivative of the complete-data log-likelihood function with respect to the parameter and solving for the parameter value at the critical point. The complete-data log-likelihood is given by
\begin{equation*}
\sum_i \log  p(\lambda^i, z^{i+1} | z^i, \theta_T, \Lambda) = \sum_i \log \mathcal{N}(\lambda^i, z^{i+1} | m, R^{-1})
\end{equation*}
with
\begin{equation*}
  R^{-1} =
  \begin{pmatrix}
    \Lambda & \Lambda A^T\\
    A\Lambda & \Omega + A\Lambda A^T
  \end{pmatrix}, \quad
  m = \begin{pmatrix}
    0\\
    z^i
    \end{pmatrix}
\end{equation*}
and
\begin{equation}
  \label{eq:deriv_op_estim_A_def}
  A_{\cdot,j} = G^j z^i.
\end{equation}
Setting the partial derivative of the complete-data log-likelihood function with respect to parameter $G$ equal to zero gives, 
\begin{equation*}
  0 =\sum_i  \frac{\partial}{\partial R} \log \mathcal{N}(\lambda^i, z^{i+1} | m, R^{-1})\frac{\partial R}{\partial A}\frac{\partial A}{\partial G}
\end{equation*}
yielding
\begin{align}
  \label{eq:partialA}
  0 &= \sum_i \frac{\partial}{\partial R} \log \mathcal{N}(\lambda^i, z^{i+1} | m, R^{-1}) \\&\propto -\sum_i\frac{\partial}{\partial R}\log|R| + \frac{\partial}{\partial R}\text{tr}(\sum_i R y y^T) \\&= -\sum_i R^{-1} + \sum_i y y^T
\end{align}
with
\begin{equation*}
  y = \begin{pmatrix}
    \lambda^i\\
    \Delta z^i
  \end{pmatrix}
\end{equation*}
where $\Delta z^i=z^{i+1}-z^i$ is the difference between sequential image representations.
The right most terms give a system of equations,
\begin{align}
  \label{eq:system}
  &\begin{pmatrix}
    \Lambda & \Lambda A^T\\
    A\Lambda & \Omega + A\Lambda A^T
  \end{pmatrix}=
                    \begin{pmatrix}
                      \lambda^i(\lambda^i)^T & \lambda^i(\Delta z^i)^T\\
                      \Delta z^i(\lambda^i)^T & \Delta z^i(\Delta z^i)^T
                      \end{pmatrix}
\end{align}
and by substitution,
\begin{equation*}
  \sum_i A \lambda^i(\lambda^i)^T = \sum_i \Delta z^i(\lambda^i)^T.
\end{equation*}
Substituting equation (\ref{eq:deriv_op_estim_A_def}) gives,
\begin{equation*}
  \sum_i G (z^i \otimes \lambda^i) (\lambda^i)^T = \sum_i \Delta z^i(\lambda^i)^T.
\end{equation*}
By a property of the Kronecker product, multiplying both sides by $(z^i \otimes \lambda^i)$ gives, 
\begin{equation*}
G = \left(\sum_i \Delta z^i(z^i\otimes \lambda^i)^T\right)\left(\sum_i z^i(z^i)^T\otimes \lambda^i (\lambda^i)^T\right)^{-1}
\end{equation*}

The update equation for $\Omega$ is derived similarly.
Taking the relevant terms in the system of equations in (\ref{eq:system}) gives,
\begin{align*}
  \sum_i A\lambda^i(\Delta z^i)^T =& \sum_i A\lambda^i(\lambda^i)^TA^T\\
  M \Omega + \sum_i A\lambda^i(\Delta z^i)^T =& \sum_i \Delta z^i(\Delta z^i)^T
\end{align*}
combining the above gives,
\begin{align*}
  \Omega = \frac{1}{M}\Big( \sum_i& \Delta z^i(\Delta z^i)^T - 2A\lambda^i(\Delta z^i)^T + A\lambda^i(\lambda^i)^TA^T\Big).
\end{align*}

\section{Derivations for PPCA (Sec \ref{sec:im_op_estim})}
\subsection{Posterior distribution of the image representation}
\label{sec:posterior_image}
The posterior distribution of the latent image representation given in equation (\ref{eq:im_op_im_rep_posterior}) is derived using the equations in Section \ref{sec:combining_gauss}. The conditional distribution of the image given
the latent image representation is a linear Gaussian distribution and the prior distribution on the latent
image representation is Gaussian. Using the equations in Section \ref{sec:combining_gauss}, the posterior distribution is given by
\begin{equation*}
  p(z^i|x^i, \theta_R) = \mathcal{N}(z^i|u, \sigma^{-2}\Sigma)
\end{equation*}
where $\theta_R = (W, \mu, \sigma)$ are the representation embedding parameters and
\begin{equation*}
  \Sigma = (I + W^TW)^{-1}, \quad u = \Sigma^{-1} W^T x^i_\mu
\end{equation*}
where $x^i_\mu=x^i - \mu$ is $x_i$ mean-centered.

\subsection{Complete-data likelihood}
The complete-data likelihood is the product of the data and latent variable distributions,
\begin{align*}
  \prod_i& p(\lambda^i, z^{i+1}, x^{i+1}, z^i| x^i, \theta_T, \Lambda, \theta_R) = \\&\prod_i p(x^{i+1} | z^{i+1}, \theta_R)\, p(z^{i+1} | z^i, \lambda^i , \theta_T, \Lambda)\, p(\lambda^i| \Lambda) \,p(z^i | x^i, \theta_R).
\end{align*}
where $\theta_T=(G,\Omega)$ are the transformation parameters.
Using the result from equation (\ref{eq:deriv_op_estim_complete_data}) 
the complete-data likelihood can be expressed
\begin{align*}
  \prod_i p(&\lambda^i, z^{i+1}, x^{i+1}, z^i | x^i, \theta_T, \Lambda, \theta_R) 
  =\\& \prod_i p(x^{i+1} | z^{i+1}, \theta_R) \,p(\lambda^i, z^{i+1} | z^i, \theta_T, \Lambda)\, p(z^i | x^i, \theta_R) 
\end{align*}
The distribution
\begin{align*}
  p(\lambda^i, z^{i+1} | z^i, \theta_T, \Lambda) &= p(z^{i+1} | z^i, \lambda^i, \theta_T, \Lambda)p(\lambda^i| \Lambda) \\&= \mathcal{N}(\lambda^i, z^{i+1} | m, R^{-1})
\end{align*}
with
\begin{equation*}
  R^{-1} =
  \begin{pmatrix}
    \Lambda & \Lambda A^T\\
    A\Lambda & \Omega + A\Lambda A^T
  \end{pmatrix}, \quad
  m = \begin{pmatrix}
    0\\
    z^i
    \end{pmatrix}
\end{equation*}
and
\begin{equation*}
A_{\cdot,j} = G^j z^i,
\end{equation*}
is given in equation (\ref{eq:deriv_op_estim_complete_data}). 

\subsection{Posterior distribution of the latent variables}
\label{sec:im_op_estim_latent}
The latent posterior distribution of the combination coefficients $\lambda^i$ and the latent image distributions $z^i$ and $z^{i+1}$ given in equation 
(\ref{eq:latent_posterior})
is derived using the equations in Section \ref{sec:combining_gauss}.

The conditional distribution of the transformed image $x^{i+1}$ given the latent representation of the transformed image $z^{i+1}$ is a linear Gaussian distribution and the conditional distribution of the latent representation of the transformed image $z^{i+1}$ given the latent representation of the initial image $z^i$ is Gaussian. Using the equations in Section \ref{sec:combining_gauss}, these distributions are combined to give the posterior distribution of the latent representation of the transformed image $z^{i+1}$ as
\begin{align*}
  p(& z^{i+1} | x^{i+1}, z^i, \lambda^i, \theta_T, \theta_R) =\\ &\frac{ p(x^{i+1} | z^{i+1}, \theta_R)p(z^{i+1} | z^i, \lambda^i, \theta_T) }{p(x^{i+1} | \theta_R)} = \mathcal{N}(z^{i+1}| \gamma, \Gamma)
\end{align*}
where
\begin{align*}
  \gamma = \Gamma \{ \sigma^{-2}W^T x^{i+1}_\mu + \Omega^{-1}(z^i + A\lambda^i)\}, \\
  \Gamma = (\Omega^{-1} + \sigma^{-2} W^T W)^{-1}, \quad A_{\cdot,m} = G^j z^i.
\end{align*}
The latent posterior distribution can then be expressed,
\begin{align*}
   p(&\lambda^i, z^{i+1}, z^i | x^{i+1}, x^i, \theta_T, \Lambda, \theta_R) = \\& p( z^{i+1} | x^{i+1}, z^i, \lambda^i, \theta_T, \theta_R )\, p(\lambda^i| \Lambda)\, p(z^i | x^i, \theta_R).
\end{align*}
This latent posterior distribution is non-Gaussian, however, 
it is still in the exponential family
and can be expressed in the canonical form,
\begin{equation*}
p(x|\eta) = h(x)\exp\{\eta(\theta)^T T(x)-A(\theta)\},
\end{equation*}
where,
\begin{align*}
  h(x) =& 3\log \frac{1}{(2\pi)^{d/2}}\\
  A(\theta) =& \Gamma\sigma^{-2}W^T  x^{i+1}_\mu(x^{i+1}_\mu)^T W\sigma^{-2} \\&+ \sigma^2\Sigma^{-2}\Sigma^{-1}W^T x^{i+1}_\mu (x^{i+1}_\mu)^T W\Sigma^{-1}\\& + \log\frac{1}{|\Gamma|^{1/2}} + \log\frac{1}{|\Lambda|^{1/2}} + \log\frac{1}{|\sigma^{-2}\Sigma|^{1/2}}\\
  \eta(\theta)= & (\Gamma^{-1}, -2\sigma^{-2}W^T x^{i+1}_\mu, -2\Omega^{-1}, -2\Omega^{-1}G^j,\\& \Omega^{-1}\Gamma\Omega^{-1} + \sigma^2\Sigma^{-1}, (G^k)^T\Omega^{-1}\Gamma\Omega^{-1}G^j,\\& 2\Omega^{-1}\Gamma\sigma^{-2}W^T x^{i+1}_\mu-2\sigma^2\Sigma^{-1}\Sigma^{-1}W^T x^i_\mu,\\& 2(G^j)^T\Omega^{-1}\Gamma\sigma^{-2}W^T x^{i+1}_\mu, 2(G^j)^T\Omega^{-1}\Gamma\Omega^{-1}, \Lambda)\\
  T(x) = & (z^{i+1}(z^{i+1})^T,  z^{i+1},  z^i(z^{i+1})^T,  \lambda^i_j z^i(z^{i+1})^T,  z^i(z^i)^T,\\&  \lambda^i_j\lambda^i_k z^i(z^i)^T,  z^i,  \lambda^i_j(z^i)^T,  \lambda^i_j z^i(z^i)^T,  \lambda^i(\lambda^i)^T)
\end{align*}

The sufficient statistics, $T(x)$, are computed by taking partial derivatives of $A(\theta)$ with respect to $\eta(\theta)$.

\subsection{Update equations for the model parameters}
The update equation for each distribution parameter is found by computing the partial derivative of the complete-data log-likelihood function with respect to the parameter and solving for the parameter value at the critical point.
The complete-data log-likelihood is given by
\begin{align*}
  \log p(&\lambda^i, z^{i+1}, x^{i+1}, z^i | \theta_T, \Lambda, \theta_R) =\\& \log \mathcal{N}(\lambda^i, z^{i+1} | m, R^{-1}) + \log \mathcal{N}(z^{i+1}|u^{i+1}, \Sigma) \\&+ \log \mathcal{N}(z^i|u^i, \Sigma)
\end{align*}
Setting the partial derivative of the complete-data log-likelihood function with respect to parameter $G$ equal to zero gives, 
\begin{equation}
  \label{eq:deriv_im_op_estim_G}
G = \left(\sum_i \Delta z^i(z^i\otimes\lambda^i)^T\right)\left(\sum_i z^i(z^i)^T\otimes \lambda^i(\lambda^i)^T \right)^{-1}
\end{equation}
where $\Delta z^i = z^{i+1} - z^i$ is the difference between sequential image representations.

Setting the partial derivative of the complete-data log-likelihood function with respect to parameter $\Omega$ equal to zero gives, 
\begin{align}
\label{eq:deriv_im_op_estim_Omega}
  \Omega = \frac{1}{M}\Big( \sum_i& \Delta z^i(\Delta z^i)^T - 2A\lambda^i(\Delta z^i)^T + A\lambda^i(\lambda^i)^TA^T\Big).
\end{align}
Update equations \ref{eq:deriv_im_op_estim_G} and \ref{eq:deriv_im_op_estim_Omega} are derived similarly to the update equations in Section \ref{sec:op_estim}.
Setting the partial derivative of the complete-data log-likelihood function with respect to parameter $W$ equal to zero gives, 
\begin{align*}
  0 = \sum_i& \frac{\partial}{\partial W}\log \mathcal{N}(x^{i+1}|W z^{i+1} + \mu, \sigma^{2}I) \\&+ \frac{\partial}{\partial W}\log \mathcal{N}(x^i|W z^i + \mu, \sigma^2I).
\end{align*}
Reordering terms gives
\begin{align*}
  W&\left(\sum_i z^{i+1}(z^{i+1})^T + z^i(z^i)^T\right) =\\ &\left(\sum_i x^{i+1}_\mu (z^{i+1})^T + x^i_\mu (z^i)^T\right),
\end{align*}
and solving for $W$ gives
\begin{align*}
  W = &\left(\sum_i x^{i+1}_\mu (z^{i+1})^T + x^i_\mu (z^i)^T\right) \\&\left(\sum_i z^{i+1}(z^{i+1})^T + z^i(z^i)^T\right)^{-1}.
\end{align*}
Setting the partial derivative of the complete-data log-likelihood function with respect to parameter $\sigma$ equal to zero gives, 
\begin{align*}
  0 =& \sum_i \frac{\partial}{\partial \sigma^{-2}}\log \mathcal{N}(x^{i+1}|W z^{i+1} + \mu, \sigma^{2}I) \\&\qquad+ \frac{\partial}{\partial \sigma^{-2}}\log \mathcal{N}(x^i|W z^i + \mu, \sigma^2I)\\
  =& \sum_i \frac{\partial}{\partial \sigma^{-2}} \log|\sigma^{-2}I| + \frac{\partial}{\partial \sigma^{-2}}\sigma^{-2}I(x^i_\mu - W z^i)(x^i_\mu - W z^i)^T \\ &\qquad+ \frac{\partial}{\partial \sigma^{-2}}\sigma^{-2}I (x^{i+1}_\mu - W z^{i+1})(x^{i+1}_\mu - W z^{i+1})^T.
\end{align*}
Reordering terms gives,
\begin{align*}
  N\sigma^2I = \sum_i&(x^i_\mu - W z^i)(x^i_\mu - W z^i u)^T \\&+ (x^{i+1}_\mu - W z^{i+1})(x^{i+1}_\mu - W z^{i+1})^T
\end{align*}
and application of the trace operator gives the update,
\begin{align*}
    \sigma^2 =& \frac{1}{ND} \text{tr}\Big(\sum_i x^i_\mu (x^i_\mu)^T + x^{i+1}_\mu (x^{i+1}_\mu)^T \\&\qquad\qquad\qquad+ W(z^i(z^i)^T + z^{i+1}(z^{i+1})^T)W^T\Big) \\&-\frac{2}{ND} \text{tr}\left(\sum_i W(z^i(x^i_\mu)^T + z^{i+1}(x^{+1}i_\mu)^T) - \mu\mu^T\right)
\end{align*}

\section{Derivations for PNPCA (Sec \ref{sec:nonlin_im_op_estim})}
\subsection{Posterior distribution of the image representation}
Computation of the posterior distribution of the latent image representation given in Section \ref{sec:nonlin_im_op_estim} is analytically intractable. A common approach in this setting is to approximate the distribution by an analytically tractable variational distribution. In Section \ref{sec:nonlin_im_op_estim} the posterior distribution of the latent image representation is approximated by the variational posterior
\begin{equation}
  \label{eq:deriv_nonlin_var_posterior}
  q(z^k|x^k, w_\phi) = \mathcal{N}(z^k | \phi(x^k)_\mu, \phi(x^k)_\sigma)
\end{equation}
where the mean and standard deviation of the distribution are functions of the encoding network parameters and the image $x^k$.
\subsection{Complete-data likelihood}
The complete-data likelihood is the product of the data and latent variable distributions,
\begin{align*}
  \prod_i &p(\lambda^i, z^{i+1}, x^{i+1}, z^i | x^i, \theta_T, \Lambda, w_\psi, w_\phi) = \\&\prod_i p(x^{i+1} | z^{i+1}, w_\psi)\, p(z^{i+1} | z^i, \lambda^i ,\theta_T)\, p(\lambda^i| \Lambda)\, p(z^i | x^i, w_\psi)
\end{align*}
Since the posterior distribution is intractable, the
variational posterior in equation (\ref{eq:deriv_nonlin_var_posterior}) is used instead. To ensure the variational
posterior is a good approximation of the true posterior distribution
requires that the KL-divergence between the variational posterior and the true posterior (see Section \ref{sec:variational_bayes}) is added to the complete-data log-likelihood function. The resulting form of the complete-data log-likelihood is
\begin{align*}
  \sum_i &\log p(\lambda^i, z^{i+1}, x^{i+1}, z^i | x^i, \theta_T, \Lambda, w_\psi, w_\phi) = \\
  &\sum_i \log p(x^{i+1} | z^{i+1}, w_\psi) + \log p(z^{i+1} | z^i, \lambda^i , \theta_T) \\ &\qquad + \log p(\lambda^i| \Lambda) + \log p(z^i | x^i, w_\psi)\\
  \geq & \sum_i \log p(x^{i+1} | z^{i+1}, w_\psi) + \log p(z^{i+1} | z^i, \lambda^i , \theta_T) \\ &\qquad +\log p(\lambda^i| \Lambda) + \mathcal{L}(q)\\
  = & \sum_i \log p(x^{i+1} | z^{i+1}, w_\psi) + \log p(z^{i+1} | z^i, \lambda^i , \theta_T) \\ &\qquad +\log p(\lambda^i| \Lambda) - \text{KL}(q(z|x, w_\phi)\,||\,p(z | w_\psi)) \\&\qquad + \mathbb{E}_{q(z|x, w_\phi)}[p(x|z, w_\psi)],
\end{align*}
where, for convenience, $p(z | w_\psi)$ is chosen to be the standard-Gaussian.

\subsection{Posterior distribution of the latent variables}
\label{sec:nonlin_op_estim_latent}
The latent posterior distribution of the combination coefficients $\lambda^i$ and the latent image distributions $z^i$ and $z^{i+1}$ given in equation 
(\ref{eq:nonlin_op_estim_latent_posterior}) is given by
\begin{align*}
   p(&\lambda^i, z^{i+1}, z^i | x^i, x^{i+1}, \theta_T, \Lambda, w_\psi, w_\phi) = \\&  \frac{ p(x^{i+1} | z^{i+1}, w_\psi)}{p(x^{i+1} | w_\psi)} p(z^{i+1} | z^i, \lambda^i, \theta_T) p(\lambda^i| \Lambda) p(z^i | x^i, w_\phi)=\\
  &  \frac{p(z^{i+1} | x^{i+1},w_\psi) }{p(z^{i+1} | w_\psi)} p(z^{i+1} | z^i, \lambda^i, \theta_T)p(\lambda^i| \Lambda) p(z^i | x^i, w_\psi)
\end{align*}
where the substitution
\begin{equation*}
  p(x^{i+1} | z^{i+1}, w_\psi) = \frac{p(z^{i+1} | x^{i+1}, w_\psi) p(x^{i+1} | w_\psi)}{p(z^{i+1} | w_\psi)}
\end{equation*}
is due to Bayes rule.
Since the posterior distribution of the latent image representation is analytically intractable, the posterior distribution of the latent variables is approximated by
\begin{align*}
   p(&\lambda^i, z^{i+1}, z^i | x^i, x^{i+1}, \theta_T, \Lambda, w_\psi, w_\psi) \approx \\&  \frac{q(z^{i+1} | x^{i+1}, w_\phi)}{p(z^{i+1} | w_\psi)} p(z^{i+1} | z^i, \lambda^i, \theta_T) p(\lambda^i| \Lambda) q(z^i | x^i, w_\phi).
\end{align*}
  This latent posterior distribution is non-Gaussian, however, 
it is still in the exponential family
and can be expressed in the canonical form,
\begin{equation*}
p(x|\eta) = h(x)\exp\{\eta(\theta)^T T(x)-A(\theta)\},
\end{equation*}
where,
\begin{align*}
  h(x) =& 3\log \frac{1}{(2\pi)^{d/2}}\\
  A(\theta) =& \text{tr}(\phi_\sigma(x^{i+1})^{-1}\phi_\mu(x^{i+1})\phi_\mu(x^{i+1})^T \\&+ \phi_\sigma(x^i)^{-1}\phi_\mu(x^i)\phi_\mu(x^i)^T) \\&+ \log\frac{1}{|\phi_\sigma(x^{i+1})|^{1/2}}+ \log\frac{1}{|\phi_\sigma(x^i)|^{1/2}} \\&+ \log\frac{1}{|\Lambda|^{1/2}} - \log\frac{1}{|I|^{1/2}}\\
  \eta(\theta) =& (\phi_\sigma(x^{i+1})^{-1} - I + \Omega,\, -2\phi_\mu(x^{i+1})\phi_\sigma(x^{i+1})^{-1},\,\\& -2\Omega^{-1},\, -2\Omega^{-1}G^j,\, \phi_\sigma(x^i)^{-1} + \Omega^{-1},\, \\&-2\Omega^{-1}G^j,\, (G^j)^T\Omega^{-1}G^k,\, \Lambda,\, -2\phi_\mu(x^i)\phi_\sigma(x^i)^{-1})\\
  T(x) =& (z^{i+1}(z^{i+1})^T,\, z^{i+1},\, z^i(z^{i+1})^T,\, \lambda^i_j z^i(z^{i+1})^T,\, z^i(z^i)^T,\,\\& \lambda^i_j z^i(z^i)^T,\, \lambda^i_j\lambda^i_k z^i(z^i)^T,\, z^i,\,
           \lambda^i(\lambda^i)^T,\, z^i)
\end{align*}
The sufficient statistics, $T(x)$, are computed by taking partial derivatives of $A(\theta)$ with respect to $\eta(\theta)$.

\subsection{Update equations for the model parameters}
The update equation for each distribution parameter is found by computing the partial derivative of the complete-data log-likelihood function with respect to the parameter and solving for the parameter value at the critical point.
The complete-data log-likelihood is given by
\begin{align*}
  \sum_i &\log p(\lambda^i, z^{i+1}, x^{i+1}, z^i | x^i, \theta_T, \Lambda, w_\psi, w_\phi) \approx \\
  &\sum_i \log p(x^{i+1} | z^{i+1}, w_\psi) + \log p(z^{i+1} | z^i, \lambda^i , \theta_T) \\ &\qquad + \log p(\lambda^i| \Lambda) + \log q(z^i | x^i, w_\phi) + \mathcal{L}(q).
\end{align*}
Setting the partial derivative of the complete-data log-likelihood function with respect to parameters $G$ and $\Omega$ equal to zero give the same results as in 
Section \ref{sec:complete_data_op} where
\begin{equation*}
G = \left(\sum_i \Delta z^i(z^i\otimes \lambda^i)^T\right)\left(\sum_i z^i(z^i)^T \otimes \lambda^i(\lambda^i)^T \right)^{-1}
\end{equation*}
and
\begin{align*}
  \Omega = \frac{1}{M}\Big( \sum_i& \Delta z^i(\Delta z^i)^T - 2A\lambda^i(\Delta z^i)^T + A\lambda^i(\lambda^i)^TA^T\Big).
\end{align*}
The parameters $w_\phi$ and $w_\psi$ are updated by backpropagation using the reparameterization trick (\cite{kingma2013auto}).

\ifCLASSOPTIONcompsoc
  \section*{Acknowledgments}
\else
  \section*{Acknowledgment}
\fi
The authors would like to thank Andrew Jaegle and Camillo J. Taylor for their insights and invaluable conversation. Support was provided by the following grants: NSF IIS 1703319, NSF
TRIPODS 1934960, NSF CPS 2038873, ARL RCTA W911NF-10-2-0016, ARL DCIST
CRA W911NF-17-2-0181, and ONR N00014-17-1-2093. 

\ifCLASSOPTIONcaptionsoff
  \newpage
\fi



%


\bibliographystyle{unsrt}
\bibliography{bibref}

%

\newpage
\begin{IEEEbiographynophoto}{Christine Allen-Blanchette}
(M '20) She received B.S. degrees in mechanical engineering and computer engineering from San Jose State University, San Jose, CA, USA in 2011, the M.S. degree in robotics from University of Pennsylvania, Philadelphia, PA, USA in 2013, and the Ph.D. degree in Computer Science from University of Pennsylvania in 2020.

She is currently a Presidential Postdoctoral Fellow at Princeton University, Princeton, NJ, USA.
\end{IEEEbiographynophoto}


\begin{IEEEbiography}[{\includegraphics[width=1in,height=1.25in,clip,keepaspectratio]{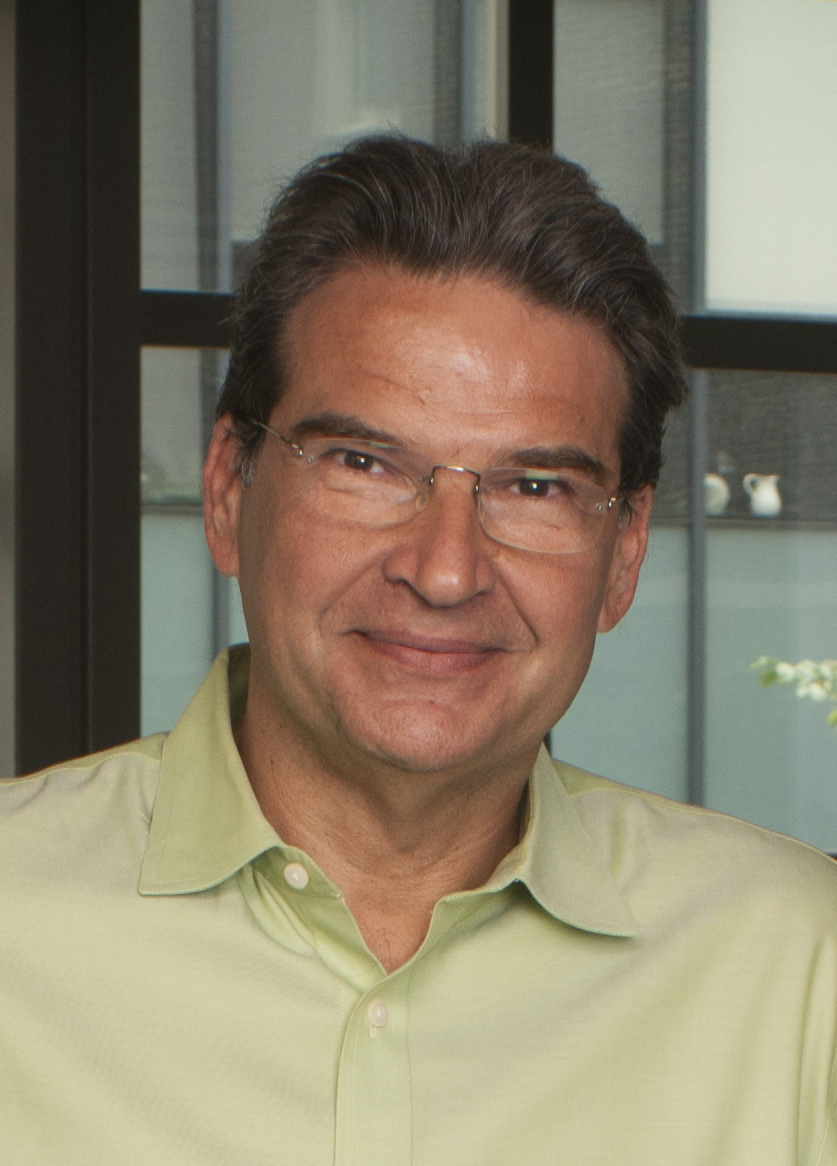}}]{Kostas Daniilidis}
Kostas Daniilidis is the Ruth Yalom Stone Professor of Computer and
Information Science at the University of Pennsylvania where he has
been faculty since 1998.  He is an IEEE Fellow.
He was the director of the GRASP laboratory from 2008 to 2013,
Associate Dean for Graduate Education from 2012-2016, and Faculty
Director of Online Learning 2012-2017. He  obtained his undergraduate
degree in Electrical Engineering from the National Technical
University of Athens, 1986, and his PhD in Computer Science from the
University of Karlsruhe, 1992.  He is co-recipient of the Best
Conference Paper Award at ICRA 2017 and Best Paper Finalist at IEEE
CASE 2015, RSS 2018, and CVPR 2019. Kostas’ main interest today is in
geometric deep learning, event-based cameras, and action
representations as applied to vision based manipulation and
navigation.
\end{IEEEbiography}


\vfill


\end{document}